\documentclass[conference]{IEEEtran}
\IEEEoverridecommandlockouts
\usepackage{cite}
\usepackage{amsmath,amssymb,amsfonts}
\usepackage{algorithm2e}
\usepackage{graphicx}
\usepackage{textcomp}
\usepackage{xcolor}
\def\BibTeX{{\rm B\kern-.05em{\sc i\kern-.025em b}\kern-.08em
    T\kern-.1667em\lower.7ex\hbox{E}\kern-.125emX}}
\begin{document}

\title{Automatic Task Parallelization of Dataflow Graphs in ML/DL models\\
}

\author{\IEEEauthorblockN{Srinjoy Das}
\IEEEauthorblockA{\textit{Univ. of Illinois at Urbana Champaign} \\
Urbana-Champaign, USA \\
srinjoy3@illinois.edu}
\and
\IEEEauthorblockN{Lawrence Rauchwerger}
\IEEEauthorblockA{\textit{Univ. of Illinois at Urbana Champaign} \\
Urbana-Champaign, USA \\
rwerger@uiuc.edu}
}

\maketitle

\begin{abstract}
Several methods exist today to accelerate Machine Learning(ML) or Deep-Learning(DL) model performance for training and inference. However, modern techniques that rely on various graph and operator parallelism methodologies rely on  search space optimizations which are costly in terms of power and hardware usage. Especially in the case of inference, when the batch size is 1 and execution is on CPUs or for power-constrained edge devices, current techniques can become costly, complicated or inapplicable. To ameliorate this, we present a Critical-Path-based Linear Clustering approach to exploit inherent parallel paths in ML dataflow graphs. Our task parallelization approach further optimizes the structure of graphs via cloning and prunes them via constant propagation and dead-code elimination. Contrary to other work, we generate readable and executable parallel Pytorch+Python code from input ML models in ONNX format via a new tool that we have built called {\bf Ramiel}. This allows us to benefit from other downstream acceleration techniques like intra-op parallelism and potentially pipeline parallelism. Our preliminary results on several ML graphs demonstrate up to 1.9$\times$ speedup over serial execution and outperform some of the current mechanisms in both compile and runtimes. Lastly, our methods are lightweight and fast enough so that they can be used effectively for power and resource-constrained devices, while still enabling downstream optimizations. 
\end{abstract}


\section{Introduction}
Deep Learning models have grown in size over the years and training or deploying them efficiently is of primary concern. Some of the well-known techniques to optimize these models involve fusion of operators \cite{fusion} while others explore techniques like TASO \cite{TASO} to automatically exploit various common algebraic transformations. As models and hardware have grown larger, the field has also moved towards improving methods to distribute models and huge training datasets across distributed systems through data, operator and pipeline parallelism also called 3D-parallelism \cite{3dparallelism}. The power and hardware resources required to train these models are enormous while search space-based methods that are usually employed to find good solutions for such device placement and distribution of compute/data require large amounts of time to obtain best-case answers. 

However, for inference with a batch size of 1, which is a frequent use case, many of the standard techniques employed today \cite{alpa,mirhoseini} that involve model/data/pipeline parallelism are inapplicable. There is no opportunity to perform parallel pipelining because only a single pipeline is executed. Opportunities for data parallelism are also completely limited if the inference is for a single datapoint. In contrast to training which requires the entire model to be run repeatedly, and backpropagate weights, a single inference run executes just once and hence the time for compiling and preparing in any auto-tuning compilers becomes important. For example, the automated ML graph optimization framework Alpa \cite{alpa} adds a 40 min compilation time overhead for optimizing training paths which should be avoided for inference. In this paper, we try to optimize the runtime of dataflow graphs used in inference for a batch size of 1 using fast automatic task-parallel algorithms. Our approach is to find structural/task-based parallelism in these graphs via clustering such that the makespan/scheduling time of these clusters is minimized, using fast algorithms.

\begin{figure}[!htb]
  \begin{center}
  \includegraphics[width=0.3\linewidth]{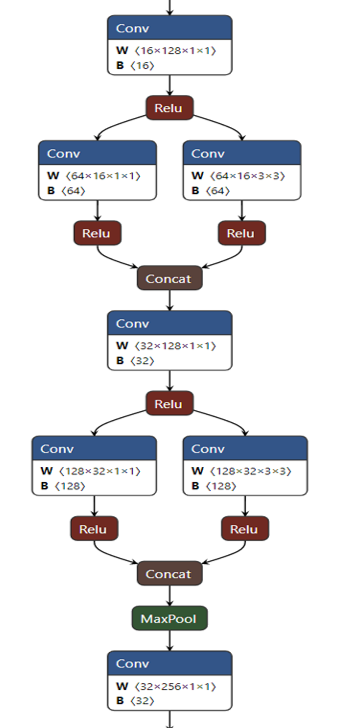}
  \end{center}
  \caption{Snippet of dataflow graph of Squeezenet, showing the Convolution, Relu and Concat operations along with the tensor dependences.}
  \label{fig:sqzsnip}
\end{figure}

 Fig.~\ref{fig:sqzsnip} shows a snippet of the dataflow graph from Squeezenet. The dataflow graph of this model shows the presence of two mutually independent paths which converge and diverge several times. We task-parallelize such graphs using path-based clustering. We can create two clusters one for each path with the cross-cluster tensor dependences being evident as messages passing across the clusters. The clusters are then scheduled on CPU-multicores and run in parallel. An important feature of our approach is that we generate parallelized high-level runnable and readable Pytorch+Python code as opposed to most of today's auto-tuning compilers for such systems. The reason being such an approach allows further downstream optimizations like intra-operator parallelism and pipeline parallelism to be applicable transparently. Our solution, though targeted for inference, may also be used as an initial optimization pass for training setups with high batch sizes. Since the goal of this work is to implement fast static compile time algorithms instead of costly search space-based methods with comparable performance profiles we resort to linear clustering algorithms. Finally, our method aims to augment the existing ML/DL dataflow graph parallelization strategies at the same time providing unique solution to the inference with batch size of 1 scenario.

\subsection{The Contributions of this Work}
\begin{itemize}
    \item Implement a recursive critical-path-based Linear Clustering algorithm augmented with a new {\bf cluster merging} step. Additionally show how to prune/optimize such graphs using task cloning and constant propagation/dead-code elimination.
    \item Design and implement a new {\bf Hyperclustering} algorithm that tries to exploit the slack times of such clusters, for batch size $>$ 1.
    \item A new tool called {\bf Ramiel } that ingests ONNX \cite{onnx} models and generates {\bf runnable} Pytorch+Python methods that can be called from Python modules - each such method representing a parallel cluster of the dataflow graph. The generated code is easily debuggable, extendable and visually readable not precluding other standard downstream optimizations.

\end{itemize}

\section{Background}
 A key challenge for ML is determining how to split a large model across multiple heterogeneous devices to achieve the fastest possible training/inference speed. Today this is typically left to human experts, but determining an optimal device placement can be very challenging, particularly as neural networks grow in complexity or approach device memory limits.
 
 {\bf Intra-Operator} (called intra-op in literature) \cite{alpa} parallelism is a method where a single ML/DL operation can be split up into several smaller operations of sub-elements in the tensor, and can be executed on multiple hardware resources- in a data-parallel fashion. Another approach called {\bf Model Parallelism} \cite{modelp} is used to break up the model into parts and distribute them on the parallel hardware. A common form of model parallelism is {\bf Pipeline Parallelism} \cite{pipelinep} where independent data samples are overlapped such that while one group of tasks execute on one set of data, another group of operations execute on a different set creating a pipeline similar to what we see in processor execution. {\bf Data Parallelism} is also popular, especially for training, where batches of training samples are sub-divided into parallel mini-batches and executed on multiple hardware resources.
 
While many of these parallelism exploitation has been done by human experts, recent developments involve auto-parallel algorithms using search-space-based or dynmaic-programming-based techniques. Some of the specific efforts that are of interest include the following. {\bf Ding et al., 2021:} IOS: In Inter-Operator Scheduler for CNN Acceleration \cite{dingios} the authors extensively study the parallelism between operators and propose Inter-Operator Scheduler (IOS) to automatically schedule multiple operators’ parallel execution through a dynamic programming algorithm.
{\bf Wang et al., 2020:} In \cite{wang} the authors propose using {\it asynchronous scheduling} for task parallelism and hyper-tuning framework parameters (like the number of cores used) via a careful study of the models.
{\bf Yi et al., 2020:} Propose FastT \cite{fastT}, a module to work with the TensorFlow framework for automatically identifying and speeding up training over multiple GPUs inspired by DAG scheduling.
{\bf Zheng et al., 2022:} Alpa is an automated model-parallel training system that generates an execution plan unifying data, operator, and pipeline parallelism \cite{alpa}. Its inter-op compilation pass distributes JAX IR into several stages, and slices the device cluster into a number of device meshes. Alpa's mechanism is costly as it involves solving integer or dynamic programming formulations and mainly targets GPU training. They beat manually tuned systems like Megatron \cite{megatron} and Deepspeed \cite{deepspeed}.
{\bf Zeng et al., 2022:}  Use dynamic critical-path-based method called FD-DPS for speeding up model training, implementing intra-operator parallelism via graph search space exploration \cite{zeng}.
{\bf Mirhoseini et al., 2017:} Propose an RNN-trained policy \cite{mirhoseini} which aims to optimize the device placement of models, but is very demanding in terms of computational power, energy costs and time.  

 \section{Methodology}

 \subsection{Observations on Structures of ML Graphs}

\subsubsection{Potential Parallelism and Observations for ML graphs}

We study several ML models - Squeezenet, Googlenet, Inception V3/V4, Yolo, BERT, Retinanet and NASNet to implement our techniques. Some of these models exhibit fork-join kind of parallel structures - for example in Squeezenet and Googlenet, while the graphs exhibit more complex structures in models like Yolo, BERT, Retinanet and NASNet. Even for fork-join graph structures the out-degree or fan-out for such fork points may vary from one region of the graph to another. 
We define the potential $Parallelism$ factor, a theoretical approximation of the parallelism that can be extracted from a dataflow graph. This factor is obtained by dividing the total amount of computation in a graph by the weighted length of the critical path (CP). More formally, \[Parallelism = Wt. Cost of Nodes/Wt. Cost of Critical Path\] where the $Wt. Cost of Nodes$ signifies an approximation for the total computation and this is calculated by applying certain static weights to the operations, heavy DL operations like $Conv$, $Matmul$ etc. having higher cost than simpler ones. Also a $Conv$ using a bigger kernel of size $7\times7$ or $5\times5$ is assigned a higher cost compared to those of size $3\times3$ or $1\times1$. Elementwise operations like $Relu$ are assigned a cost of 1. Similarly $Wt. Cost of Critical Path$ is calculated by considering the operations lying on the critical path of the graph. We also add a unit cost for each graph edge when computing the CP signifying some amount of overhead for the tensor dependences. It is due to this that for smaller graphs with long dependency chains, the potential parallelism may turn out to be $<$ 1.

\begin{figure}[!htb]
  \begin{center}
  \includegraphics[width=0.7\linewidth]{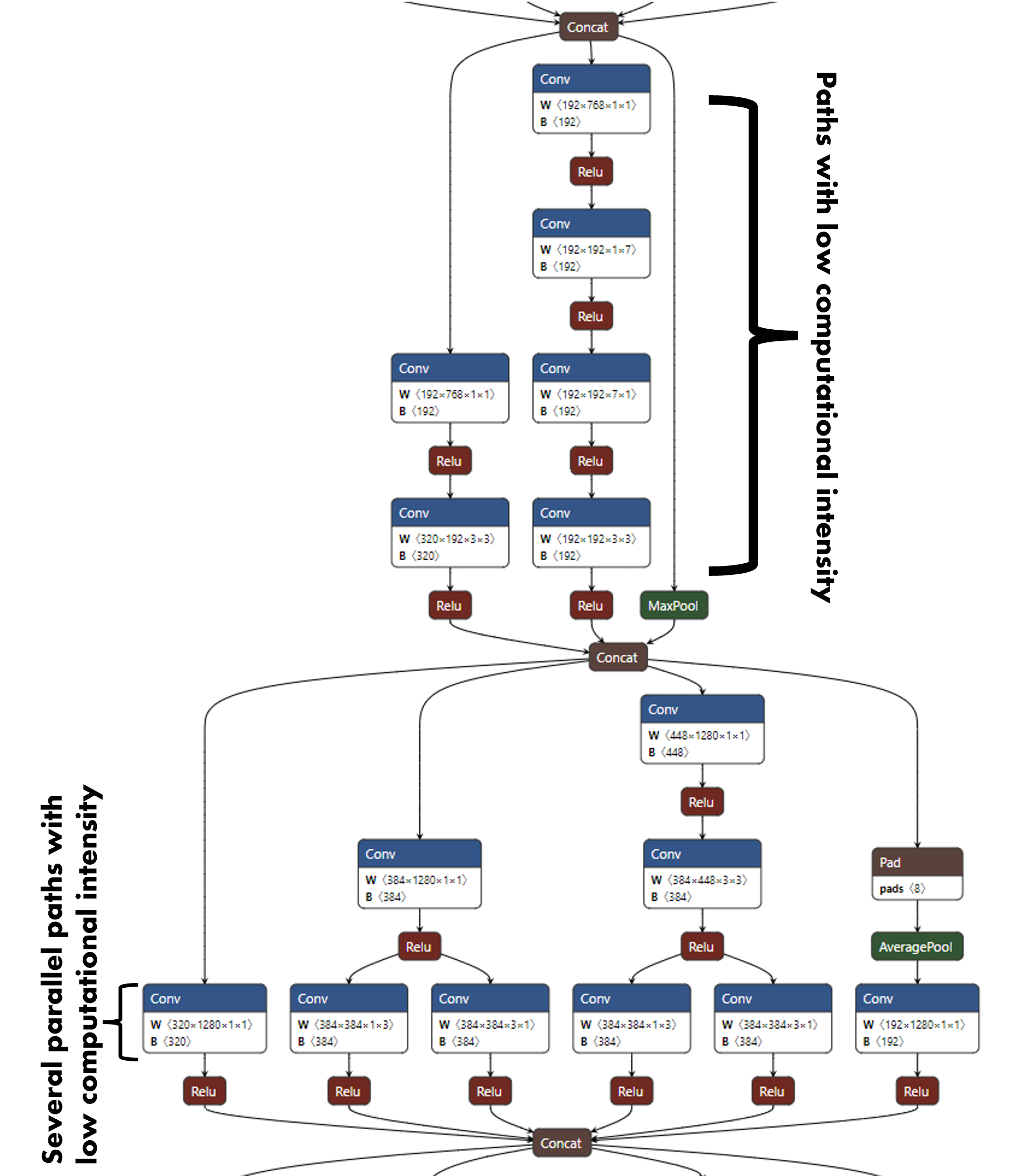}
  \end{center}
  \caption{Inception V3/V4 dataflow graph with paths/regions having low computational complexity that may reduce the parallelism exploitation opportunity.}
  \label{Fig:iv3}
\end{figure}

Looking at Inception V3/V4 as shown in Fig.~\ref{Fig:iv3} we see that some parallel paths have very low computational intensity - implying that if these paths are scheduled on independent CPU cores, those cores may be idle most of the time. This observation leads to some strategies for parallelism exploitation that we have used, like task cloning and hyperclustering.

\begin{figure}[!htb]
  \begin{center}
  \includegraphics[width=0.6\linewidth]{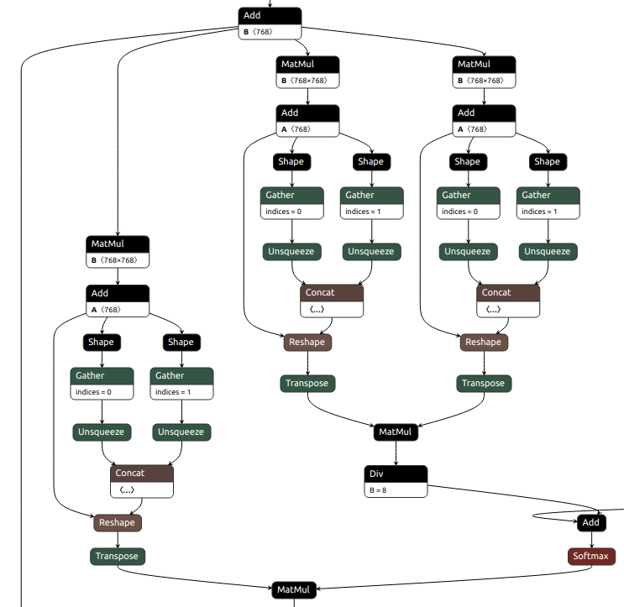}
  \end{center}
  \caption{BERT: This is a dataflow graph from one of the popular Transformer models, where we see that a certain kind of structure that arises repeatedly due to the Multi-Headed Attention mechanism that is the central premise of the model.}
  \label{Fig:bert}
\end{figure}

BERT on the other hand (shown in Fig.~\ref{Fig:bert}) demonstrates a more complex graph structure where a bunch of nodes corresponding to the multi-headed attention (MHA) sequence \cite{bert}, hang off one node - this structure being repeated many times, depending on the size of the Transformer module. The MHA sub-graph structure lends itself to further optimizations and simplification via constant propagation and dead-code elimination. We see similar opportunities for graph pruning in Yolo V5 also.

\begin{figure}[!htb]
  \begin{center}
  \includegraphics[width=1.2\linewidth]{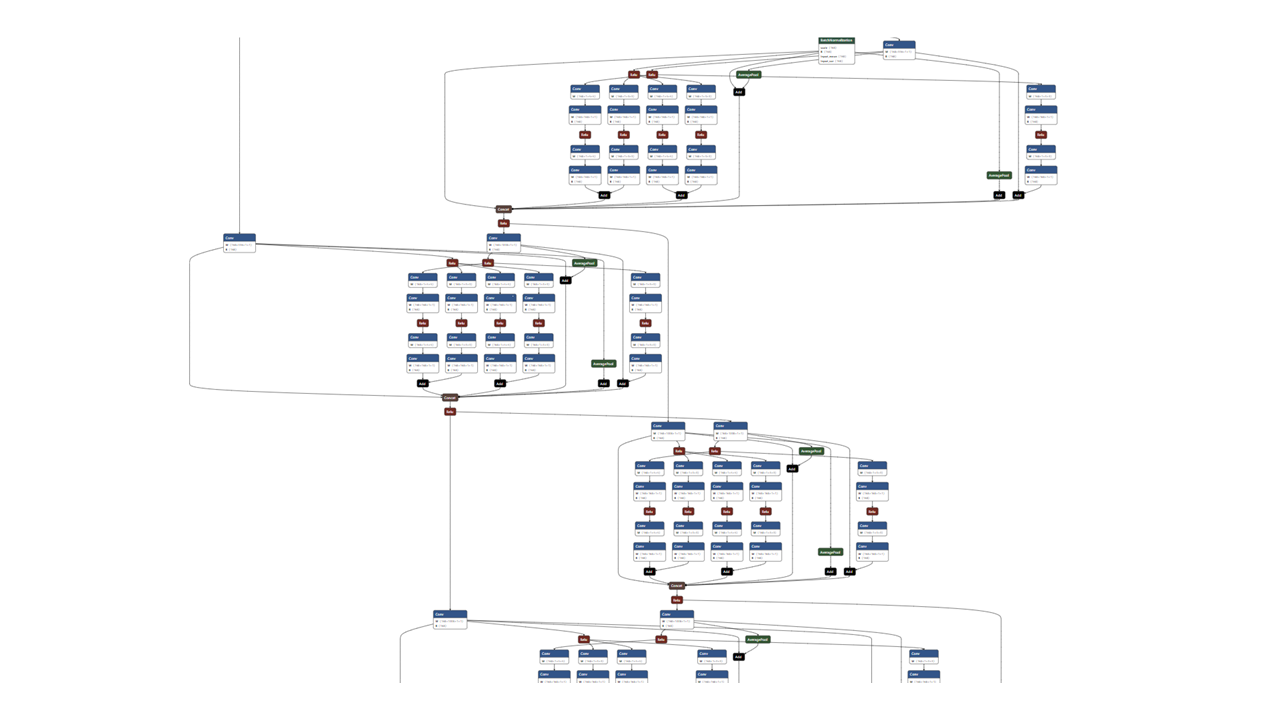}
  \end{center}
  \caption{NASNet: A neural architectural search model whose dataflow graph is big and exhibits abundant parallelism.}
  \label{Fig:nasnet}
\end{figure}

One of the biggest and most complex models we have handled is NASNet. NASNet belongs to the family of Neural Architectural Search models \cite{nasnet} and its dataflow graph contains more than a thousand nodes and edges. It has a huge fan-out at certain parts due to the search nature of the model. The graph is a mix of computationally heavy Conv2D nodes and simpler operations like slice, gather and reshape. There is abundant parallelism throughout the graph which leads to high speedup potential. A part of the structure is shown in Fig.~\ref{Fig:nasnet}. NASNet also lends itself to substantial graph pruning.

\begin{table}[!htb]
\centering
\small
\caption{\label{tab:parallelism} Potential Parallelism that exists in ML Dataflow Graphs.} 
\begin{tabular}{|l r  r r r|} 
 \hline
 Model Name  &  \#Nodes & Wt. NodeCost & Wt. CP & $\|ism$ \\ 
 \hline\hline
 Squeezenet  &  66 & 187 & 218 & 0.86$\times$ \\ 
 \hline
 Googlenet &  153 & 373 & 264 & 1.4$\times$ \\
 \hline
 Inception V3 & 238 & 1136 & 829  & 1.37$\times$ \\
\hline
 Inception V4 &  339 & 1763 & 1334 & 1.32$\times$ \\
\hline
 Yolo V5&  280 & 730 & 619 & 1.18$\times$ \\
\hline
 Retinanet &  450 & 1291 &  1102 & 1.2$\times$ \\
\hline
 BERT &  963 & 21357 & 16870 & 1.27$\times$ \\
\hline
 NASNet &  1426 & 8147 & 2187 & 3.7$\times$ \\
\hline

\end{tabular}
\end{table}

\subsubsection {Some dataflow graph metrics}

Table~\ref{tab:parallelism} represents some metrics of the dataflow graphs. We observe that some of the dataflow graphs exhibit a limited amount of parallelism. For example, in Squeezenet, the potential speedup is 0.86$\times$ which may likely cause slowdown when run on hardware. Inception V3/V4 etc. on the other hand demonstrate parallelism potential of around 1.3$\times$ to 1.4$\times$. Finally there are models like NASNet that exhibit potential parallelism of the order of 3.7$\times$. We would like to use this static parallelism metric as an indicator of actual speedup on hardware when we parallelize the models.

 \subsection{Task Parallelization  via Linear Clustering (LC) }

Clustering and scheduling a dataflow graph involves splitting the graph into a set of nodes, where each set would be a target for a different piece of hardware viz. a core of a CPU. 
Given a task graph, the {\bf clustering} problem can be defined as the process of mapping the nodes of graphs onto labelled clusters $C_0, C_1, ...,C_k$ such that the running time of the parallel code is minimized. All the tasks that belong to the same cluster execute on the same core while the clusters themselves are run on separate cores communicating via messages. 

The Recursive Critical-Path-based Linear Clustering(LC) technique by Kim et al. \cite{kimlc} was chosen as the clustering algorithm. It involves finding a critical path at the outset. Once a critical path is found, its nodes are removed from contention, and a new iteration begins to find the next critical path through the remainder graph. These iterations continue until there are no more paths available.  What we obtain is a clustering of the graph where each cluster is the longest path obtainable at that instance of the graph and is linear in nature.
The passes in our LC algorithm are the following:  i) Graph creation pass - Converts an input ONNX model into an internal representation ii) Distance pass - Computes the weighted distance of each node from the end node of the graph and stores in $distance\_to\_end$ iii) Critical path-based clustering pass and iv) Cluster merging pass. Details on last two passes follow.

\subsubsection{Critical Path-Based Clustering pass}
\begin{figure}[!htb]
  \begin{center}
  \includegraphics[width=1.0\linewidth]{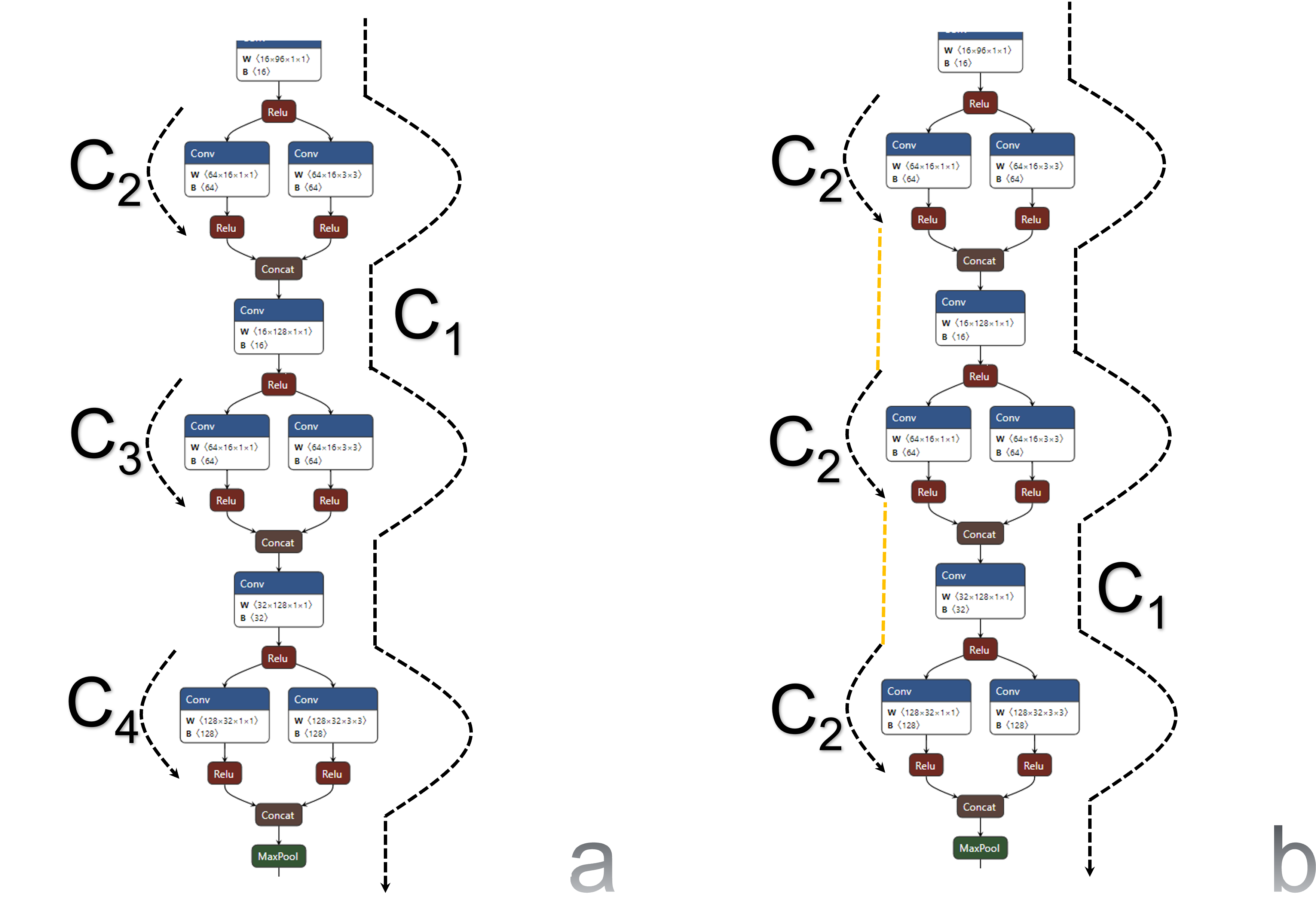}
  \end{center}
  \caption{a. Shows the clusters $C_1,\ldots, C_4$. b. Several side clusters $C_2$, $C_3$ and $C_4$ are merged to form a single cluster $C_2$.}
  \label{Fig:clustermerging}
\end{figure}

We use the critical path algorithm recursively to linearly cluster the entire graph. Algorithm 1 contains the details.
Fig.~\ref{Fig:clustermerging} shows how the clusters look when LC is applied to Squeezenet's dataflow graph. While the main cluster $C_1$ running on one core consists of heavier {\bf Conv2D$\rightarrow$Relu$\rightarrow$Concat} ops pattern repeated multiple times, the side clusters $C_2$, $C_3$ and $C_4$ are just small clusters consisting of lighter {\bf Conv2D$\rightarrow$Relu} ops pattern. To avoid multiple small clusters we apply a cluster merging pass post-LC, to merge {\bf non-overlapping clusters}. An example of how the clusters look after clustering (a) before merging and  (b) after merging is given in Fig.~\ref{Fig:clustermerging}.

\begin{algorithm}[!htb]
\small
\SetKwInOut{Input}{Input}\SetKwInOut{Output}{Output}

\Input{Dataflow Graph - $Graph$}
\Output{A clustering of the nodes of Graph, $C_1, \ldots ,C_k$, s.t. $C_i \bigcap C_j == \phi$ and every node appears in some $C_i$}
\BlankLine

nodes $\leftarrow$ $Graph.nodes$ \; 
outgoing\_edges $\leftarrow$ Outgoing edges of nodes \;
incoming\_edges $\leftarrow$ Incoming edges of nodes \;
\BlankLine 

allClusters $\leftarrow$ \{ \}\;

\While{nodes $ \neq \phi$ }{
\# {\bf Start a new Critical Path} \;
\BlankLine

readyL $\leftarrow$ List of Nodes s.t. indegree of node is 0 \;
cNode $\leftarrow$ $argmax_n$(distance\_to\_end(n)), n $\in$ readyL \; 
newCluster $\leftarrow$ \{cNode\} \;
nodes $\leftarrow$ nodes - \{cNode\} \;
\BlankLine 

\While{outgoing\_edges(cNode) $\neq \phi$ }{
succ $\leftarrow$ successor nodes of cNode \;
sNode $\leftarrow$ $argmax_n$(distance\_to\_end(n)), n $\in$ succ \;
newCluster $\leftarrow$ newCluster $\bigcup$ \{sNode\} \;
nodes $\leftarrow$ nodes - \{sNode\} \;
Remove all outgoing\_edges(cNode) other than cNode $\rightarrow$ sNode from outgoing\_edges\;
Remove all incoming\_edges(sNode) from incoming\_edges\;
cNode $\leftarrow$ sNode \;
}
\# {\bf newCluster contains the Critical Path} \;
allClusters $\leftarrow$ allClusters $\bigcup$ newCluster\; 
} 
\Return{$allClusters$}\;
\BlankLine 
\caption{Recursive Linear Clustering}\label{criticalcluster}
\end{algorithm}

\subsubsection{Cluster Merging pass}
Linear clustering has some drawbacks when applied to the ML dataflow graphs.
Due to the nature of these graphs, zeroing out the nodes in the critical path leaves behind a disconnected graph in several cases. Also, the paths generated become shorter than the previous paths because disjointedness prevents them from fully extending. The cluster merging algorithm (Algorithm 2 and 3) remedies this by iteratively combining clusters that do not overlap with respect to their start/end times and is computationally linear in the number of clusters already created. Algorithm 2 is called iteratively in Algorithm 3 till a fixed point is reached and no further merging is possible. Table~\ref{tab:clustermerging} shows the number of clusters created by the original Linear Clustering pass compared to the number of clusters finally produced by the cluster merging pass. For several models the number of linear clusters decreases significantly thereby helping to manage the parallel code generation and subsequent overheads.

\begin{algorithm}
\small
\SetKwInOut{Input}{Input}\SetKwInOut{Output}{Output}

\Input{Clusters - $C_1, \ldots C_k$}
\Output{$a.$ The merged clusters $MC_1, \ldots MC_m$, $m \leq k$, \ 
$b.$ Boolean flag mergeDone }
\BlankLine

MergedClusters $\leftarrow$ \{\}; skipCluster $\leftarrow$ \{\}\;
mergeDone $\leftarrow$ False \;

\ForEach{cluster $cl_1$ in clusters}{
\ForEach{cluster $cl_2$ in clusters}{
\BlankLine

\If{ $cl_1 \neq cl_2$ $\wedge$ ($cl_1$ and $cl_2 \notin$ skipCluster) } {
sSpan of a cluster $\leftarrow$ distance\_to\_end(entry\_node(cluster)) \;
eSpan of a cluster $\leftarrow$ distance\_to\_end(exit\_node(cluster)) \;

\# {\bf Check that cluster spans do not overlap } \;
\If{$sSpan(cl_1) < eSpan(cl_2) \parallel$ \ 
    $sSpan(cl_2) < eSpan(cl_1) $} {
Create a merged cluster $MC_i = cl_1 \bigcup cl_2$ \;
MergedClusters $\leftarrow$ MergedClusters $\bigcup MC_i$ \;
\# Do not process $cl_1$ and $cl_2$ further  \;
\# So add them to skipCluster \;
\BlankLine

add $cl_1$ and $cl_2$ to skipCluster\;
mergeDone $\leftarrow$ True ; break ;
}
}
}
\# if $cl_1$ is not merged with any cluster \;
\# {\bf Add $cl_1$ as a standalone merged cluster} \;
\If{ $cl_1$ not Merged }{
MergedClusters $\leftarrow$ MergedClusters $\bigcup cl_1$ \;
}
}
\Return{MergedClusters, mergeDone} \;
\BlankLine

\caption{MergeClusters}\label{clustermerging}
\end{algorithm}

\begin{algorithm}[!htb]
\small
\SetKwInOut{Input}{Input}\SetKwInOut{Output}{Output}

\Input{Clusters -  $C_1, \ldots C_k$}
\Output{The final merged clusters $MC_1, \ldots MC_m$, $m \leq k$}
\BlankLine

nClus $\leftarrow$ Clusters\;
mergeDone $\leftarrow$ True \;
\While{mergeDone == True }{
    MergedClusters,mergeDone $\leftarrow$ MergeClusters(nClus) \;
    nClus $\leftarrow$ MergedClusters\;

}
\Return{MergedClusters} \;
\BlankLine

\caption{Iterative Cluster Merging}\label{iterclustermerging}
\end{algorithm}

\begin{table}[!htb]
\centering
\caption{\label{tab:clustermerging} Number of clusters formed for the ML graphs, before and after {\bf Cluster Merging}.}
\begin{tabular}{||l r r||} 
 \hline
 Model  & Before Merging & After Merging\\
 \hline\hline
 Squeezenet  &  9 &  2\\ 
 \hline
 Googlenet &  30 &  4\\
 \hline
 Inception V3 &  38 &  6\\
\hline
 Inception V4 &  55 &  6 \\
\hline
 Yolo V5 &  29 &  12 \\
\hline
 BERT &  76 &  5\\
\hline
 Retinanet &  16 &  10\\
\hline
 NASNet &  244 &  67 \\
\hline
\end{tabular}
\end{table}

\subsection{Dataflow Graph Pruning via Constant Propagation (CP) and Dead-Code Elimination (DCE)}

\begin{figure}[!htb]
  \begin{center}
  \includegraphics[width=0.9\linewidth]{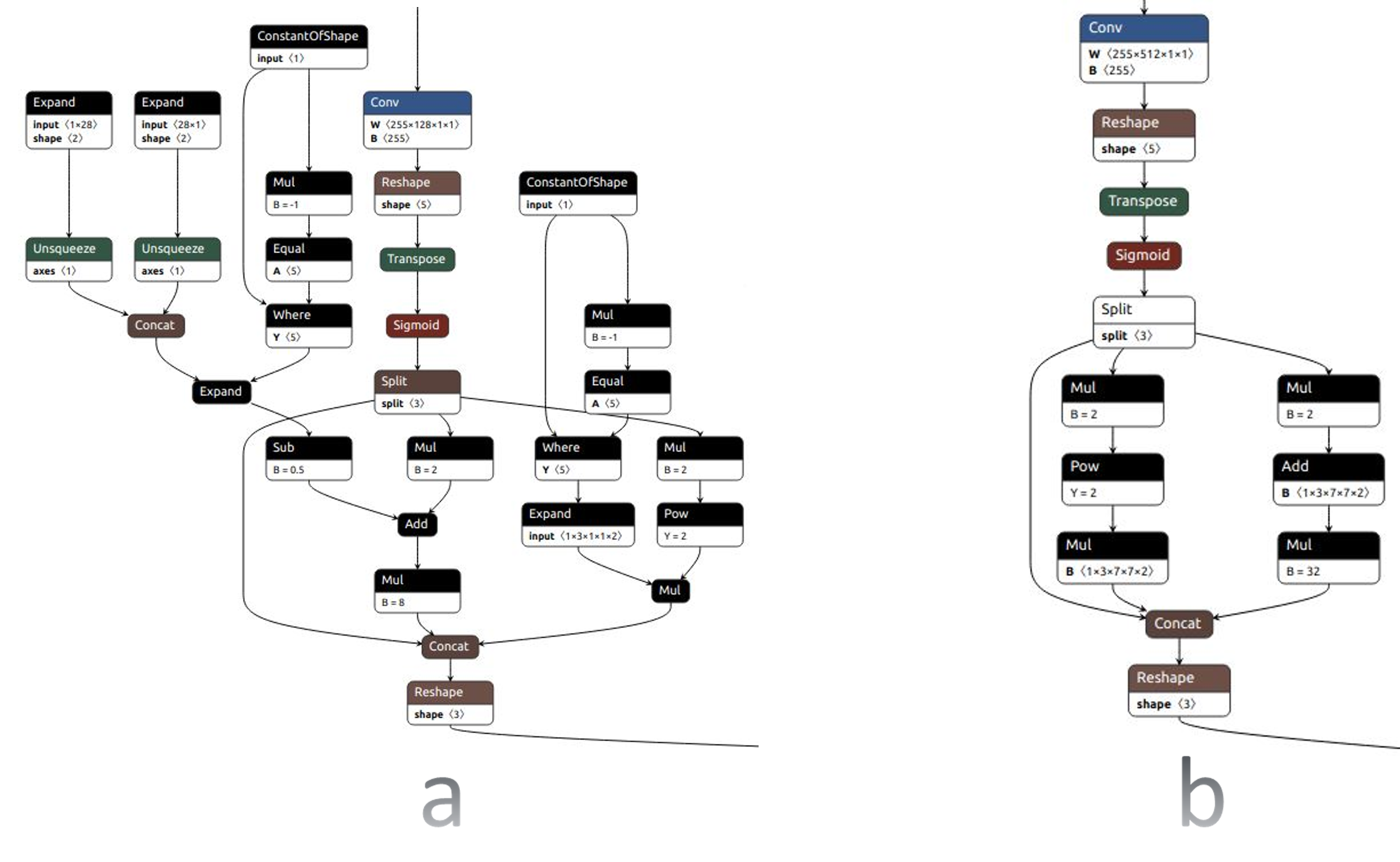}
  \end{center}
  \caption{Parts of Model Yolo before and after constant propagation.}
  \label{Fig:cprop}
\end{figure}

Constant propagation and folding eliminate constants that are used as part of the model. Further, Dead Code elimination can remove these operator nodes. We leverage ONNX Runtime ({\bf onnxruntime}) \cite{onnxruntime} to perform this graph transformation.
Viewing the simplified graph in Fig.~\ref{Fig:cprop}(b) clearly shows how the complexity and number of paths through just one section of the graph reduce from Fig.~\ref{Fig:cprop}(a). If the Cluster Merging Pass is viewed as a {\bf Vertical branch compression} strategy, then constant propagation is a {\bf Horizontal branch reduction} strategy. 
A table comparing the number of parallel clusters generated by onnxruntime before and after the optimization on the graph is shown in Table~\ref{tab:DCE}.

\begin{table}[!htb]
\centering
\caption{\label{tab:DCE} Cluster Size post Constant Propagation and Dead-Code Elimination.}
\begin{tabular}{||l r r||} 
 \hline
 Model Name &  Before Const Prop & After Const Prop  \\
 \hline\hline
 YOLO V5 & 12 & 9 \\ 
 \hline
 NASNet & 67 & 9 \\ 
 \hline
 BERT & 5 & 3 \\ 
 \hline
\end{tabular}

\end{table}

\subsection{Graph Cloning}

\begin{figure}[!htb]
  \begin{center}
  \includegraphics[width=0.85\linewidth]{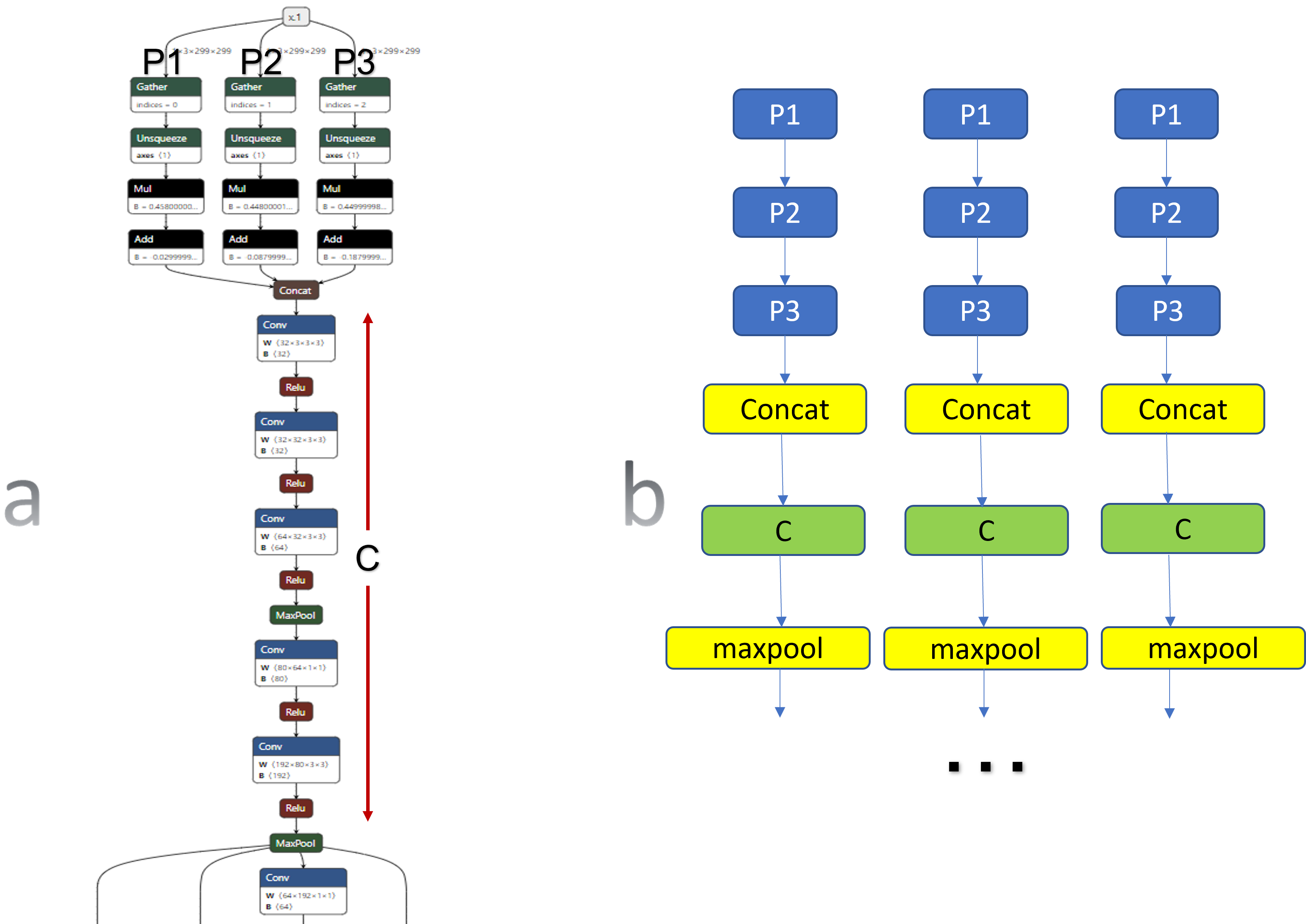}
  \end{center}
  \caption{Cloning in Inception V3.}
  \label{Fig:iv3clone}
\end{figure}

Graph Cloning is a well-studied optimization in parallel computing whereby graph nodes are cloned or replicated across multiple processors or hardware so that the overall schedule time or latency of executing the graph is reduced \cite{cloning}. Cloning is usually employed in distributed message-passing scenarios to overcome communication bottlenecks but we have applied it to our case and observed improved runtimes. Though cloning usually results in improved performance it comes at the price of $redundant$ computation happening on multiple clusters. Also, cloning can blow up the size of the graph exponentially. Hence, it should be applied with care and in limited setting. Cloning in Inception V3 dataflow graph is shown in Fig.~\ref{Fig:iv3clone}.

\subsection{Hyperclustering: Beyond Linear Clustering}

\begin{figure}[!htb]
  \begin{center}
  \includegraphics[width=0.9\linewidth]{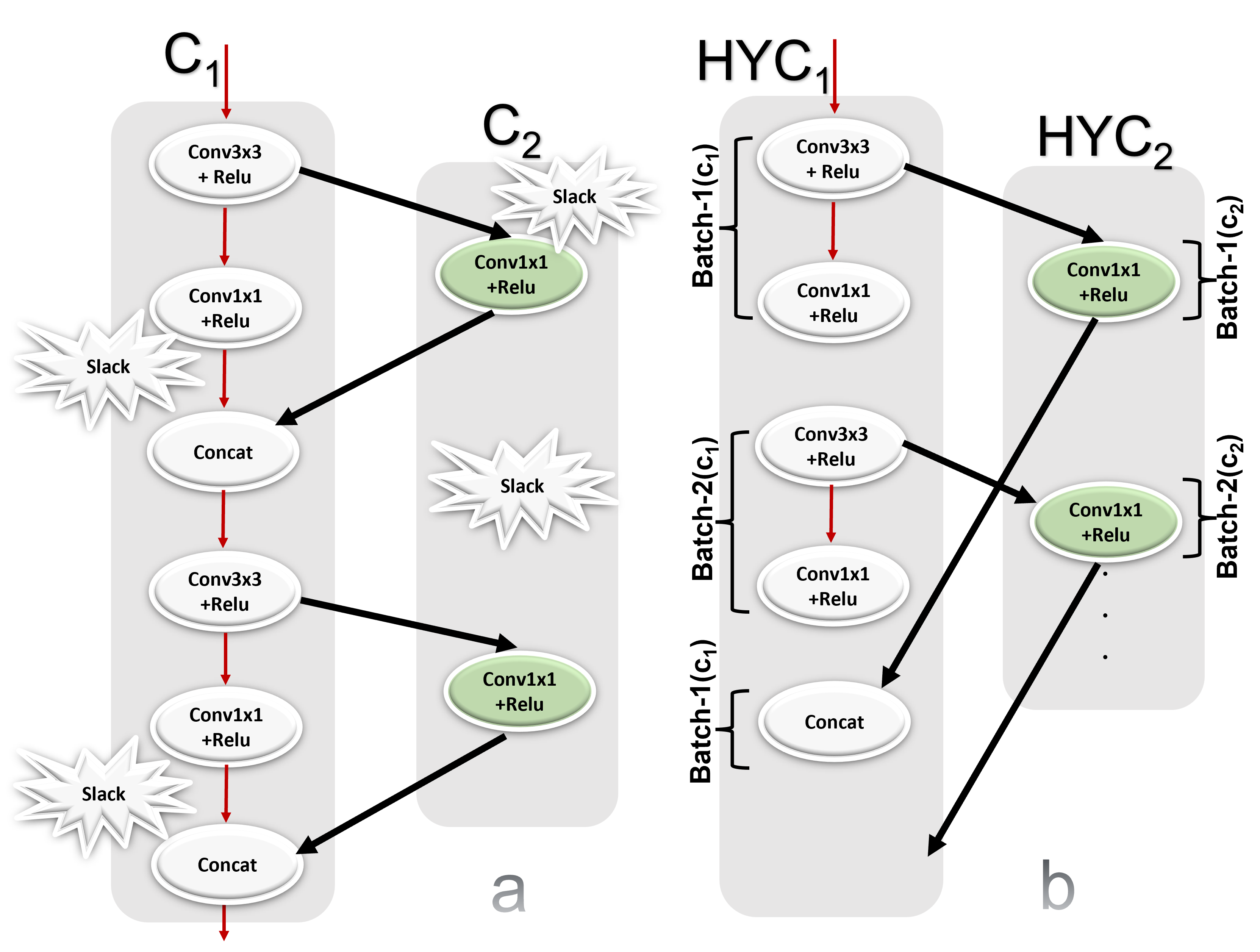}
  \end{center}
  \caption{The creation of hyperclusters is shown for Squeezenet. (a) shows the original clustering and b) shows the hyperclusters - $HYC_1$ and $HYC_2$ being created with copies from two images from a batch size of 2.}
  \label{Fig:hypercluster}
\end{figure}

When we run some profiling on the parallel code that we generate it is seen that every time a cluster waits to receive data from another cluster there arises a slack or gap especially if the data is not ready in another cluster. This imbalance is a window of opportunity if inference is done with small batch sizes $>$ 1. Taking ideas from {\bf hyperthreading} as applied to modern CPU cores, we have multiple inference samples in flight which is possible when the batch size for inference is greater than 1. This helps in reducing the slacks. Fig.~\ref{Fig:hypercluster} shows the hyperclusters created for a batch size of 2 for Squeezenet. This technique can be also applied for higher batch sizes which have the advantage of reducing and probably eliminating a large part of the slack. 



\begin{figure}[!htb]
  \begin{center}
  \includegraphics[width=0.5\linewidth]{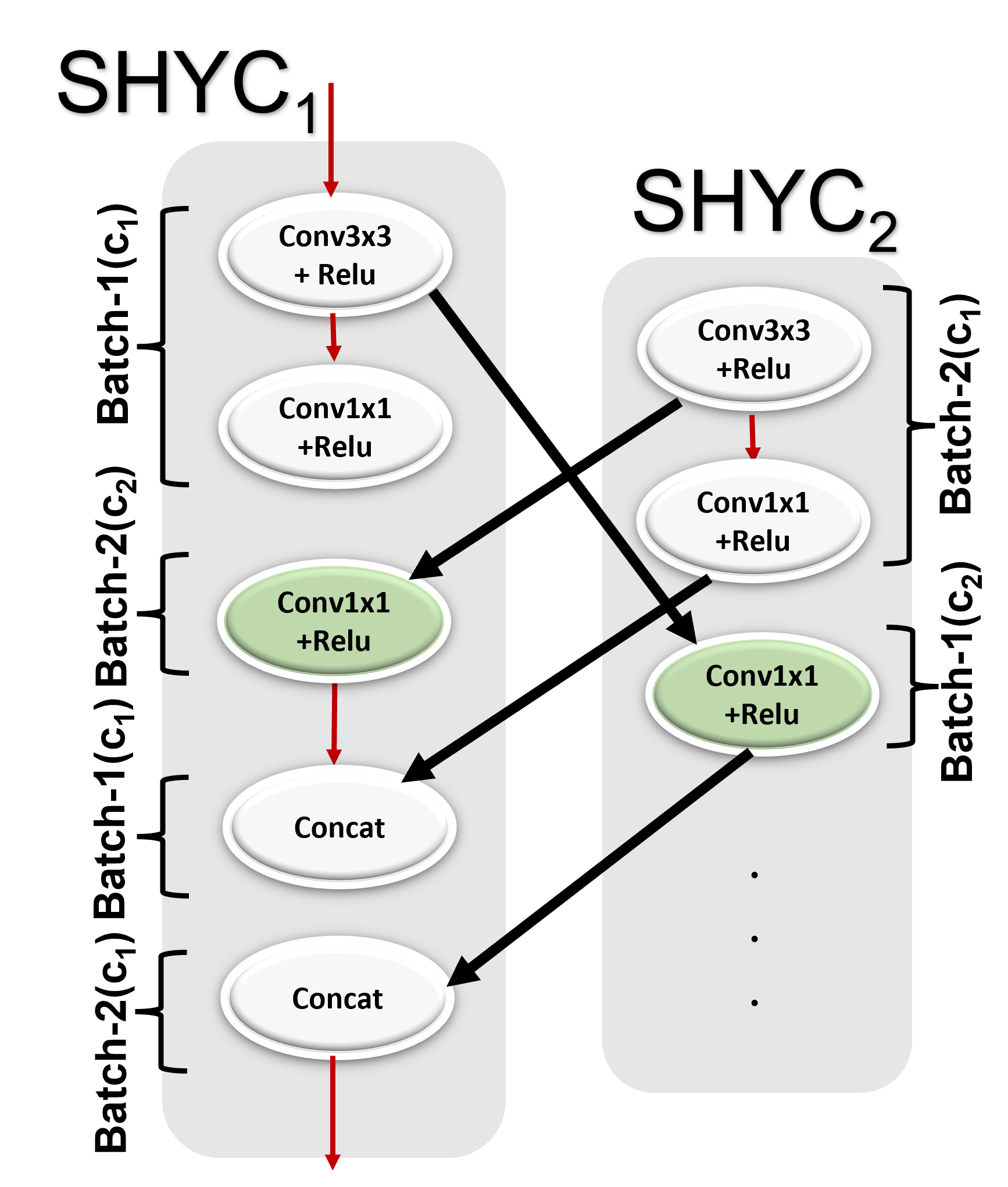}
  \end{center}
  \caption{Switched hypercluster for Squeezenet.}
  \label{Fig:switched}
\end{figure}

A more efficient method of creating hyperclusters is by aligning operations from multiple batches of dissimilar clusters so that better load balance is achieved. For each set of operations of batch$_i$, the next set of operations will be those of batch$_j$, but instead of picking operations from the same cluster, it uses the operations from one of the other clusters such that the clusters now have more balanced operational efficiency. This approach is outlined in Fig.~\ref{Fig:switched} where for the same Squeezenet example, we switch the cluster operations while creating the new hyperclusters. It can be observed that compared to hyperclustering we now have a better load balanced cluster, as we have 5 operations in $SHYC_1$ and 3 operations in $SHYC_2$ versus 5 and 2 in $HYC_1$ and $HYC_2$. We refer to such Hyperclusters as {\bf Switched Hyperclusters}.
Implementing switched hyperclusters automatically is complex for more involved ML graphs. At present we handle this automatically for simple graphs like Squeezenet and  manually for more complex graphs after observing the profile data.

 \section{Implementation Details}

\begin{figure}[!htb]
  \begin{center}
  \includegraphics[width=1.0\linewidth]{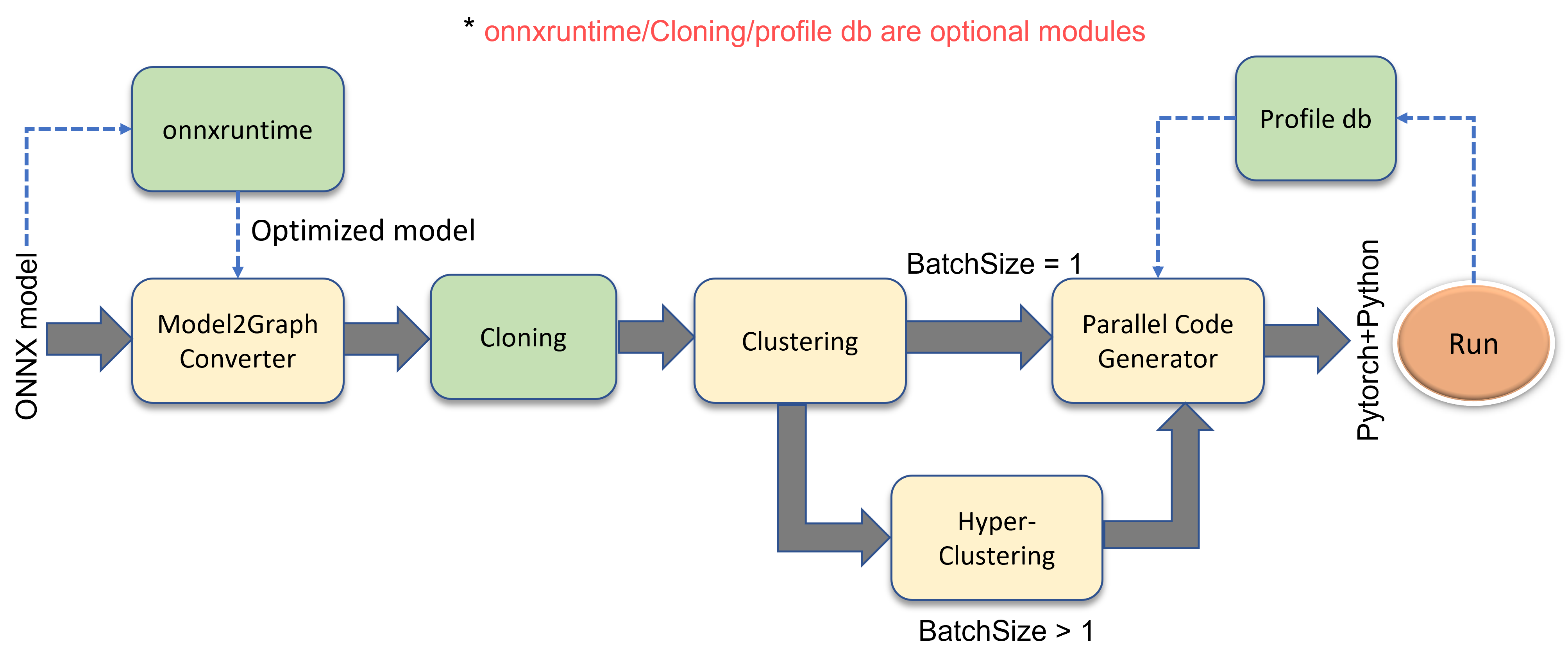}
  \end{center}
  \caption{Architecture of {\bf Ramiel}.}
  \label{Fig:ramiel}
\end{figure}

For our e2e implementation we built a new tool called {\bf Ramiel}. The high-level architecture of Ramiel is provided in Fig.~\ref{Fig:ramiel}. Ramiel's inputs are ONNX models. These models are extracted from Pytorch 2.0 repository \cite{pytorch2.0} and Huggingface \cite{huggingface} as well as ONNX model zoo \cite{onnxmodels}. ONNX provides a mechanism to load and read the frozen graph representations of the ML models via various APIs. 
Ingesting ONNX models for runnable Pytorch+Python code generation is non-trivial. Also, the conventions used in ONNX to represent the ML/DL ops are sometimes significantly different from their avatars in Pytorch. In this tool, the $Model2Graph$ Convertor stage is responsible for taking an ONNX graph and convert it to an internal in-memory graph format. onnxruntime may be used as a separate plugin at the input stage to perform constant propagation and dead code elimination and return an optimized ONNX model that is fed to this converter. An optional $Cloning$ stage is used to do limited cloning of graphs if posssible. Next, the $Clustering$ stage is responsible for performing the linear clustering algorithm and cluster merging. When the batch size is $>$ 1, the $Hyperclustering$ stage is triggered and the operations in different clusters are interleaved. Finally, $Parallel Code generator$ is responsible for converting the graph into its corresponding PyTorch+Python code. When profiling is enabled a profile database holds information about the execution trace and the slacks during communication which can be used offline.

\begin{algorithm}[!htb]
\small
\SetKwInOut{Input}{Input}\SetKwInOut{Output}{Output}

\Input{ Clusters – $C_1$ … $C_m$}
\Output{Pytorch + Python code for each cluster $C_i$, encapsulated as a method}
\BlankLine
\ForEach{cluster $C_i$}
{
Generate cluster $C_i$’s Python method definition and signature with name and input/output tensors\;
\ForEach{node n in $C_i$}{
\ForEach{each successor node succn}{
\If{succn belongs to a remote cluster $C_j$}{
Generate a send message using queue.put()\;
}
}
\ForEach{predecessor node predn}{
\If{predn belongs to a remote cluster $C_j$}{Generate a recv message using queue.get()\;}
}
Generate a new SSA-name for the output variable of the node n called outVar\;
{\bf GeneratePytorchCodeForOperandType(n)}\;
}
}

\caption{Parallel Code Generation}\label{parallelcode}
\end{algorithm}

\begin{figure*}[!htb]
  \begin{center}
  \includegraphics[width=0.8\linewidth]{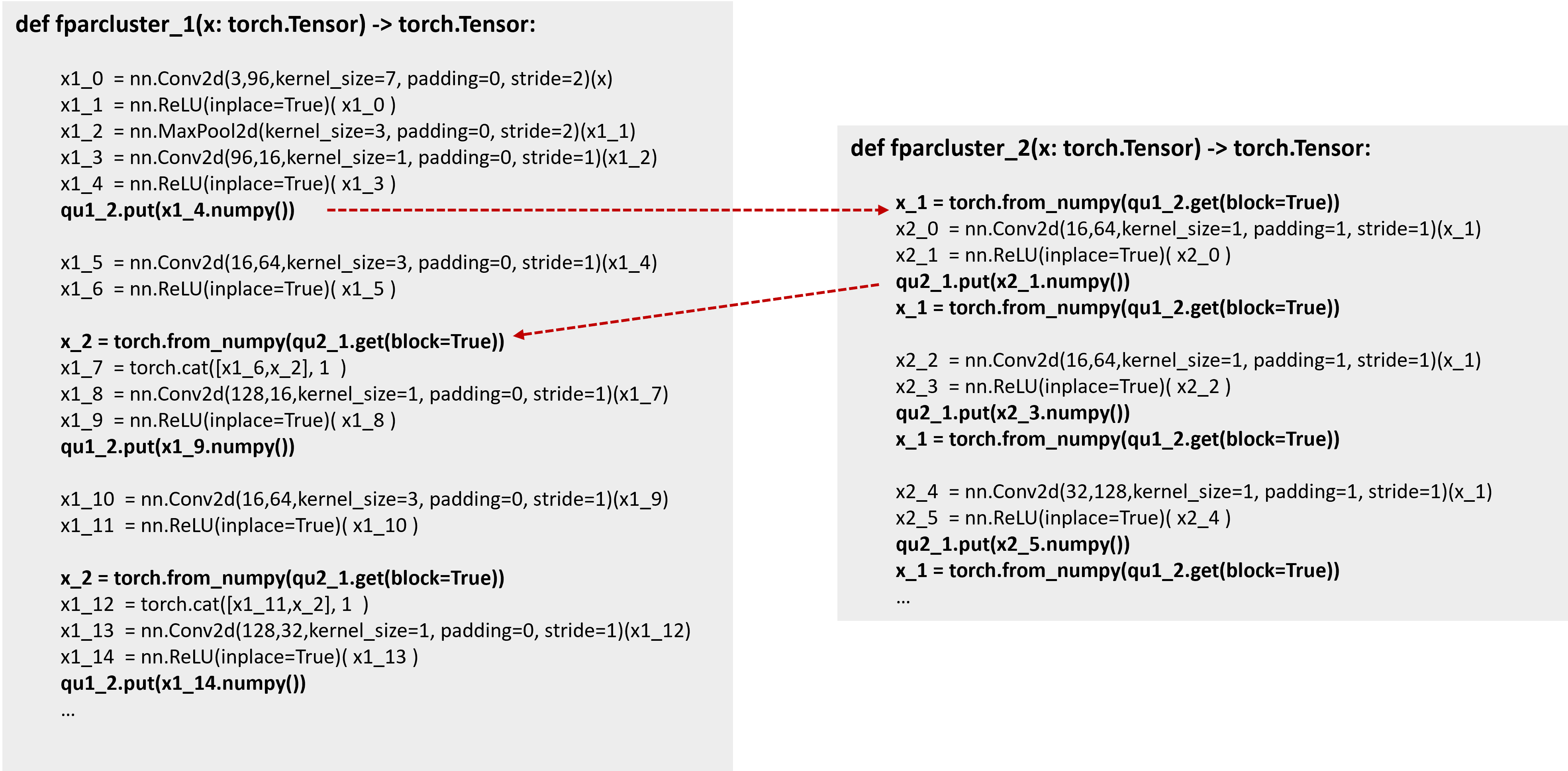}
  \end{center}
  \caption{Snippet of Parallel Code Generated for Clusters of Squeezenet using Pytorch+Python. The inter-cluster communication primitives are highlighted.}
  \label{Fig:parallelcode}
\end{figure*} 

Since our system is built on Python and Pytorch, we step away from threads due to the limitations of the Global Interpreter Lock \cite{GIL} and use Python processes. Algorithm 4 provides an overview of the parallel code generation logic.
The inputs to the parallel code generation pass are the merged clusters. Each cluster/path gets its own function, where such functions are finally forked by Python processes. Our algorithm also generates all the synchronizing {\bf get/put} primitives using the message passing bi-directional queues. {\bf GeneratePytorchCodeForOperandType} is a method which calls the equivalent Pytorch operator.
A snippet of a 2-cluster parallel code of Squeezenet is provided in Fig.~\ref{Fig:parallelcode}. To ensure completeness and to evaluate the parallel code generation, a single core non-parallel version of the code is also generated by Ramiel.

\section{Experimental Results}
Our implementation is written in Python 3.8 and Pytorch is used as the ML framework. The parallel code that is generated by the tool Ramiel also targets PyTorch and Python 3.8. An Intel 12-core 24-thread 1.1Ghz Xeon Gold machine was used for all runtime experiments. {\bf Other than what is specified for hyperclustering experiments, all other experiments have been conducted with a batch size of 1}. The experiments target ML dataflow graphs for inference setups. 

\begin{table*}[!htb]
\centering
\caption{\label{tab:lc} Performance of Linear Clustering (LC) algorithm.}
\small
\begin{tabular}{||l r r r r r||} 
 \hline
 Model Name &  Parallelism & No. of clusters & Seq Time(ms) & Parallel Time(ms) & Speedup \\
 \hline\hline
 Squeezenet & 0.86$\times$ & 2 & 85 & 103 & 0.83$\times$ \\ 
 \hline
 GoogleNet & 1.4$\times$ & 4 & 188 & 156 & 1.2$\times$ \\
 \hline
 Inception V3 & 1.37$\times$ & 6 & 559 & 422 & 1.32$\times$\\
 \hline
 Inception V4 & 1.32$\times$ & 6 & 1212 & 840 & 1.44$\times$\\
 \hline
 Yolo V5 & 1.18$\times$ & 12 & 790  & 820 & 0.96$\times$\\ 
 \hline
 BERT & 1.27$\times$ & 6 & 3296 & 3071 &  1.07$\times$\\
 \hline
 Retinanet & 1.2$\times$ & 10 & 4311 & 3361 & 1.3$\times$\\ 
 \hline
 NASNet & 3.7$\times$ & 67 & 2271 & 1351 & 1.7$\times$\\
 \hline
\end{tabular}
\end{table*}

\noindent
\subsection{Performance of Linear Clustering (LC)}
\noindent
Our first set of experiments shown in Table~\ref{tab:lc} are execution time comparisons between sequential and parallel code generated by the Linear Clustering (LC) algorithm after cluster merging. This does not take into account further optimizations such as constant propagation, cloning or hyperclustering. Both the sequential and parallel code are automatically generated by Ramiel. 
The results in this table signify a direct correlation to the potential parallelism factor. Squeezenet demonstrates a parallelism factor of 0.86, a value lower than 1 indicating that it is infeasible to attempt to parallelize this. When the parallelism factor rises higher than 1 there is a clear trend of speedup in runtime. NASNet shows the highest speedup of 1.7$\times$ for LC though the potential parallelism stands much higher at 3.7$\times$. This is likely due to the large number of clusters (67 of them) created by LC and the potential overhead of scheduling and running them on Pytorch. However, with advanced optimizations the number of clusters decreases substantially pushing this performance higher.
For other graphs like Inception V3/V4 the achieved parallelism is close to the projected values. Yolo and BERT have low projected parallelism. Hence we  achieve a slight drop in performance with LC for Yolo and only a moderate 1.07$\times$ with BERT. Retinanet beats the static parallelism estimate by achieving 1.3$\times$ speedup.

\noindent
\subsection{Performance of LC + Downstream Intra-Op Parallelism}
\noindent
We enable Intra-op parallelism as a downstream optimization post-LC, as shown in Table~\ref{tab:lc_intra} and we compare our LC+Intra-op runtimes with pure intra-op runtimes instead of comparing with sequential runs. In Pytorch, the operations are parallelized (intra-op parallelism) using OpenMP \cite{openmp} constructs provided by Pytorch. Hence, by changing the number of OpenMP threads used, we can vary the degree of intra-op parallelism. We enable OpenMP threads in Pytorch using the Intel OpenMP library (libiomp) \cite{libiomp}. Enabling intra-op parallelism allows operations such as a single convolution to use multiple threads. We show that using a combination of LC + downstream intra-op parallelism helps get better speedup than pure intra-op parallelism or LC in most of the cases. Some of the models demonstrate slight slowdowns or plateauing with higher openmp threads, which is due to oversubscription of the CPU cores.

\begin{table*}[!htb]
\centering
\caption{\label{tab:lc_intra} Performance of Linear Clustering (LC) + Downstream Intra-Op Parallelism ({\bf * Both Par and Seq have intra-op enabled}).}
\small

\begin{tabular}{||c||c c c|c c c|c||}
 \hline
 Model Name & \multicolumn{3}{c} {NUM\_THREADS=2} & \multicolumn{3}{c}{NUM\_THREADS=4}  & Overall Speedup \\ 
 \hline
 & Par(ms) & Seq(ms) & Speedup & Par(ms) & Seq(ms) & Speedup & Best Overall = Best Seq Time/Best Par Time \\
 \hline
 Squeezenet & 86 & 67 & 0.78$\times$ & 87 & 58 &  0.67$\times$ & 0.67$\times$ \\ 
 \hline
 Googlenet & 143 & 163 & 1.14$\times$ & 143 & 143 & 1.00$\times$ & 1.00$\times$\\
 \hline
 Inception V3 & 375 & 476 & 1.27$\times$ & 357 & 438 & 1.23$\times$ & 1.23$\times$\\
 \hline
 Inception V4 & 676 & 983 & 1.45$\times$ & 699 & 826 & 1.18$\times$ &  1.22$\times$\\
 \hline
 Retinanet & 2100 & 2575 & 1.23$\times$ & 1462 & 1633 & 1.12$\times$ & 1.12$\times$ \\
 \hline
NASNet & 1297 & 1664 & 1.3$\times$ & - & - & - & 1.3$\times$ \\
 \hline
\end{tabular}
\end{table*}

\begin{table}[!htb]
\centering
\caption{\label{tab:lc_dce} Performance of LC when augmented with Constant Propagation and Dead-Code Elimination.}
\small
\begin{tabular}{||l r r||} 
 \hline
 Model  & S$_{LC}$  & S$_{LC+DCE}$\\
 \hline\hline
 Yolo V5  &  0.96$\times$ &  1.06$\times$\\ 
 \hline
 BERT &  1.07$\times$ &  1.15$\times$\\
 \hline
 NASNet &  1.7$\times$ & 1.91$\times$ \\
 \hline
\end{tabular}

\end{table}

\noindent

\subsection{Performance of LC + Constant Propagation(CP) + Dead-code Elimination(DCE)}
\noindent
When Constant Propagation and Dead Code Elimination are applied via {\bf onnxruntime} as shown in Table~\ref{tab:lc_dce} (here $S_{LC}$ and $S_{LC+DCE}$ stand for respective speedups), we see gains for those graphs where constant propagation prunes the graphs. Squeeznet, Googlenet and Inception models do not demonstrate the presence of constants. However, Yolo, NASNet and BERT present several opportunities to carry out dead-code elimination.
Yolo had an apparent slowdown with a 4\% loss in performance with LC when not using constant propagation and subsequently shows a performance improvement with a close to 10\% upswing in speedup. 
NASNet also gains well with an 1.9$\times$ uplift when compared to 1.7$\times$ without CP and DCE. Overall, CP+DCE seems to be an essential optimization to prune ML/DL dataflow graphs for improved clustering and runtimes. 

\noindent
\subsection{Performance of LC + Cloning}

\begin{figure}[!htb]
  \begin{center}
  \includegraphics[width=0.9\linewidth]{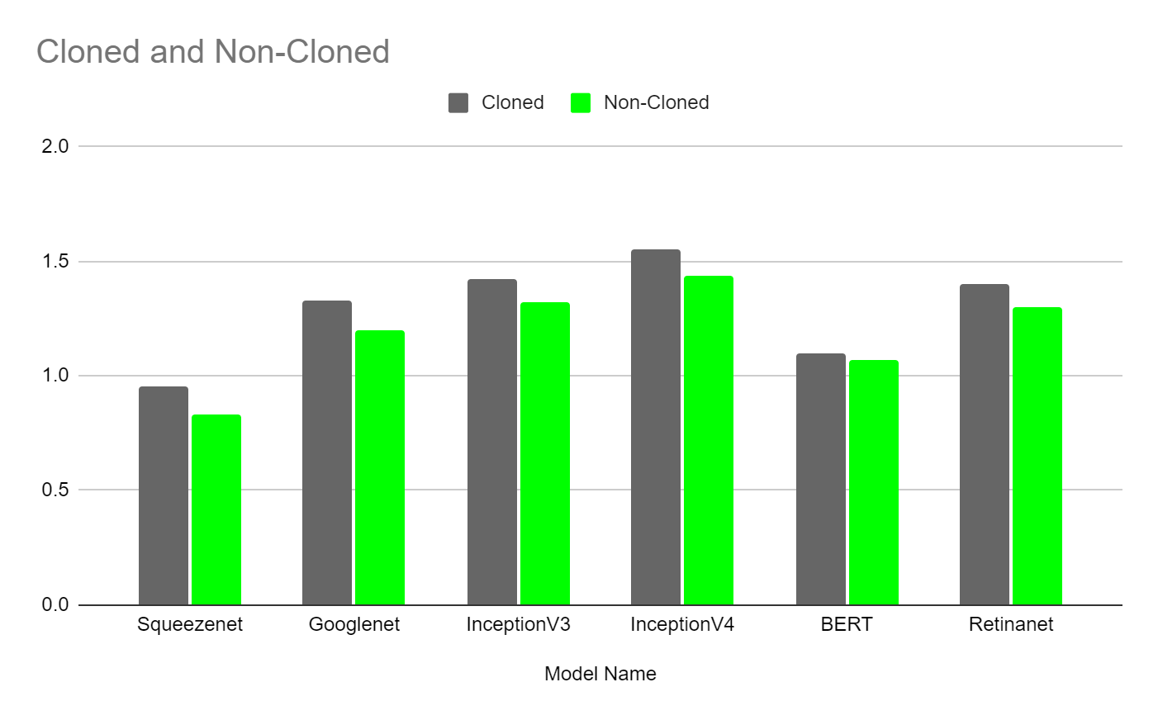}
  \end{center}
  \caption{Performance uplift of cloned models versus non-cloned models.}
  \label{Fig:cloned_perf}
\end{figure}

\noindent
Restricted cloning has been applied in our work, mostly at the top half of the dataflow graphs to reduce the blow-up in terms of size. For our experiments, we have cloned the smaller dataflow graphs and avoided bigger graphs like NASNet. Cloning provides moderate boost in performance with up to 8\% improvement. Details are in Fig.~\ref{Fig:cloned_perf}.

\noindent

\subsection{Overall Performance and Comparison with Inter-Operator Scheduler(IOS)}
\noindent

\begin{table}[!htb]
\centering
\caption{\label{tab:lc_clone} Performance of LC when augmented with CP+DCE+Cloning.}
\small
\begin{tabular}{||l r r r r||} 
 \hline
 Model Name &  S$_{LC}$ & S$_{LC+DCE}$ & S$_{LC+Cloning}$ & S$_{Overall}$\\
 \hline\hline
 Squeezenet & 0.83$\times$ & - & 0.95$\times$ & 0.95$\times$\\ 
 \hline
 GoogleNet & 1.2$\times$ & - & 1.33$\times$ & 1.33$\times$ \\
 \hline
 Inception V3 & 1.32$\times$ & - & 1.42$\times$ & 1.42$\times$\\
 \hline
 Inception V4 & 1.44$\times$ & - & 1.55$\times$ & 1.55$\times$ \\
 \hline
 BERT & 1.07$\times$ & 1.15$\times$ & 1.1$\times$ & 1.18$\times$ \\
 \hline
Yolo V5 & 0.96$\times$ & 1.06$\times$ & - & 1.06$\times$ \\
 \hline
Retinanet & 1.3$\times$ & - & 1.4$\times$ & 1.4$\times$ \\
 \hline
NASNet & 1.7$\times$ & 1.91$\times$ & - & 1.91$\times$ \\
 \hline
\end{tabular}
\end{table}


\noindent
In Table ~\ref{tab:lc_clone} we show the overall impact of LC, CP+DCE, and cloning when compared to sequential implementation. 
We also compare our inference speedup with the work done by Ding et al. \cite{dingios} named IOS. In IOS the authors extract inter-operator parallelism for small batch sizes using a dynamic-programming algorithm. Hence, this work is closest to ours as we also extract inter-operator/task parallelism for small batch sizes. Table~\ref{tab:comp} shows the comparison of speedup attained by our work (Speedup$_{Ours}$) compared to IOS - Speedup$_{IOS}$, for benchmarks that are in common between our experimental studies. These are Squeezenet, Inception and NASNet. CT(s) stands for compile-time in seconds. Our work (Table~\ref{tab:comp}) is much faster than IOS with respect to runtimes for NASNet (1.91$\times$ vs 1.4$\times$), on par for Inception (1.6$\times$) and moderately slower on Squeezenet (which is due to higher Python process overheads in our implementation). On the compile time front IOS is significantly costlier than our mechanism. Even for large graphs like NASNet/BERT/Yolo etc. we complete our code generation in a few seconds - NASNet consuming the highest time of 9.7 seconds. On the other hand, IOS takes about 90 minutes to generate the schedule for NASNet and around a minute each for Squeezenet and Inception. Hence, our mechanism is faster by 10$\times$ to 500$\times$ with similar to better runtimes, thereby saving on important parameters like power and cost.

\begin{table}[!htb]
\centering
\caption{\label{tab:comp} Performance Comparison of our work with IOS.}
\begin{tabular}{||l r r r r r ||} 
 \hline
 Model Name &  Speedup$_{Ours}$ & CT(s)& Speedup$_{IOS}$ & CT(s) & \\
 \hline\hline
 Squeezenet & 0.95$\times$ & 2.2s & 1.15$\times$ & 60s &\\ 
 \hline
 Inception & 1.55$\times$ & 5.2s & 1.59$\times$ & 60s & \\
 \hline
NASNet & 1.91$\times$ &  9.7s & 1.4$\times$ & 5400s & \\
 \hline
\end{tabular}

\end{table}

\noindent
\subsection{Performance of Hyperclustering}

\begin{figure*}[htb]
  \begin{center}
  \includegraphics[width=0.65\linewidth]{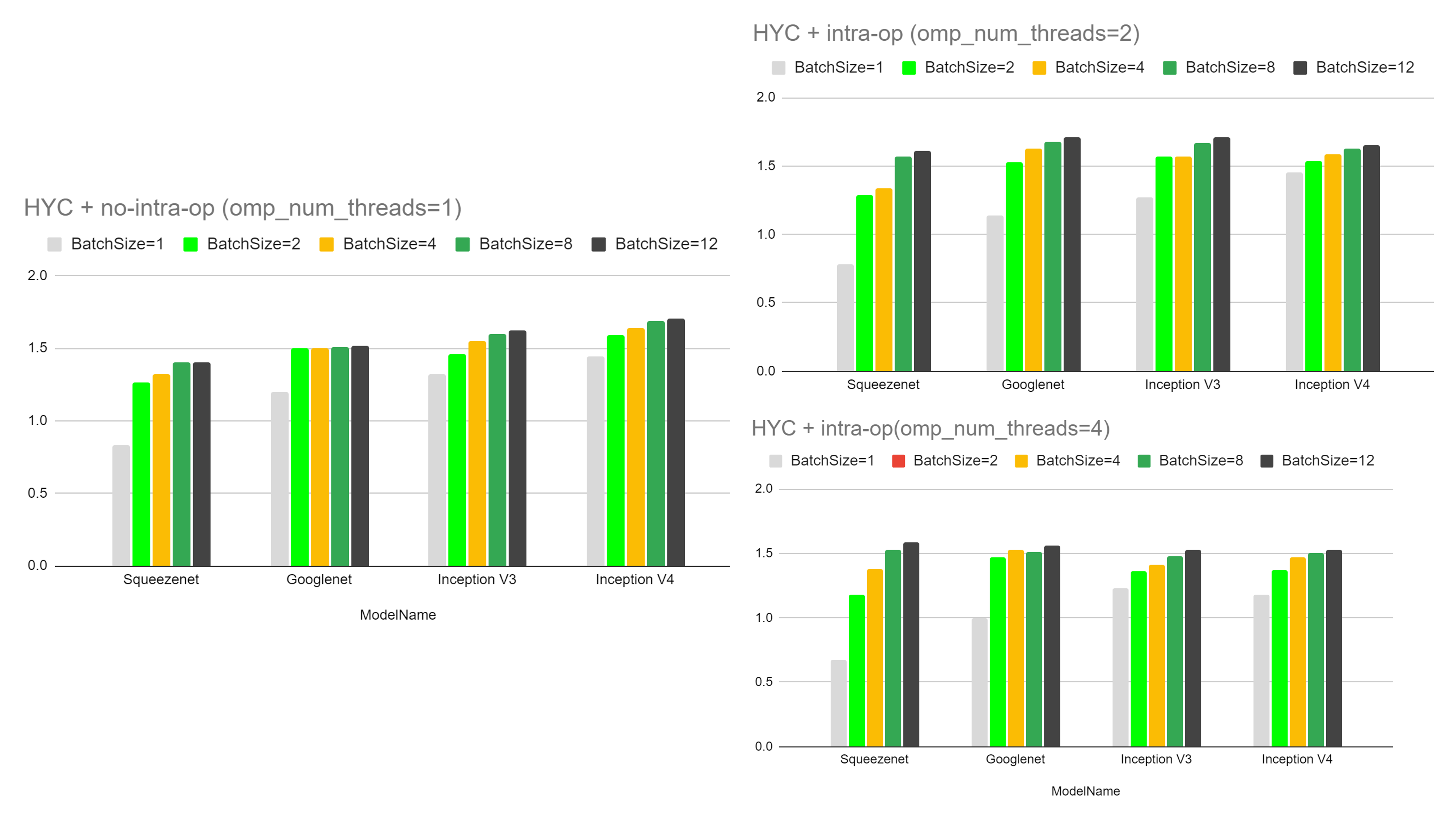}
  \end{center}
  \caption{Performance of Hyperclustering with batch sizes of 2, 4, 8, 12 with and without intra-op.}
  \label{Fig:hyc}
\end{figure*}

\begin{figure*}[!htb]
  \begin{center}
  \includegraphics[width=0.55\linewidth]{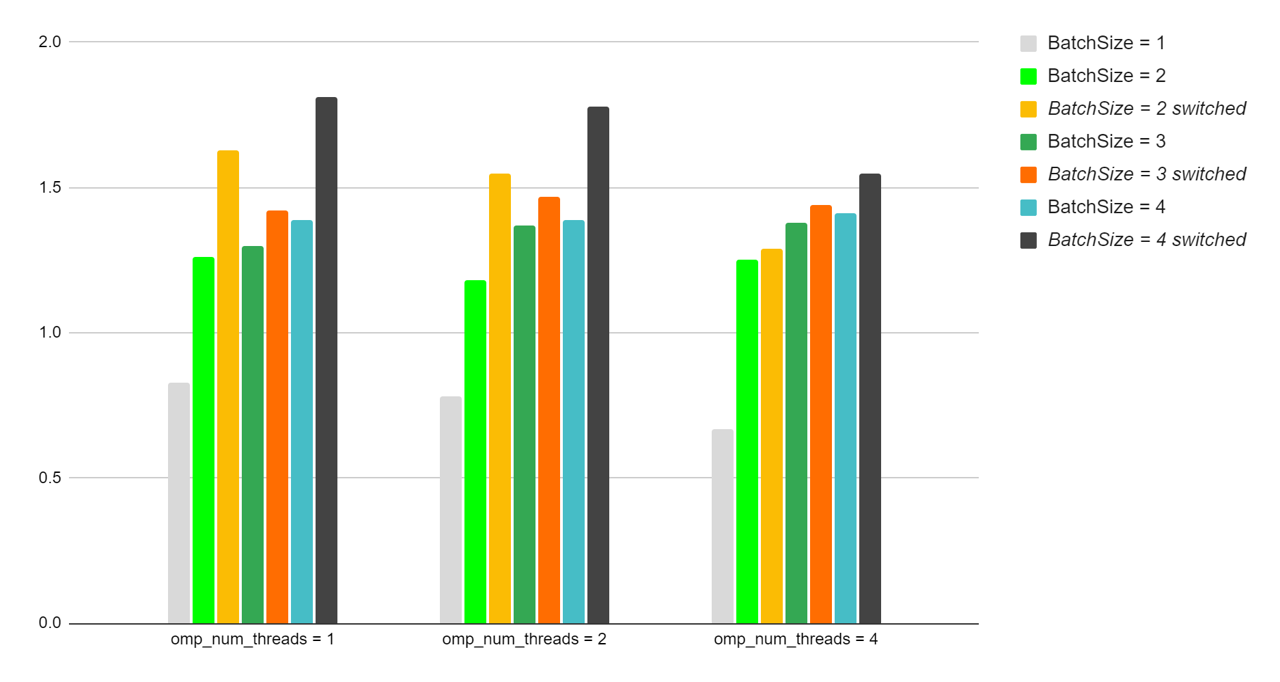}
  \end{center}
  \caption{Performance of Switched Hyperclustering with batch sizes of 2, 3, 4 with and without intra-op for Squeezenet}
  \label{Fig:swhyc}
\end{figure*}

For Hyperclustering, we use batch sizes of 2, 4, 8 and 12 and observe the relative speedup when compared to the sequential version of the codes. As evident from Fig.~\ref{Fig:hyc}, the potential speedup exploitable when Hyperclustering is used are significant. The speedup  rises as we increase the number of threads (up to hardware thread limit). The switched Hyperclustering algorithm further improves performance when compared to the non-switched versions of the code with performance uplifts of around 30\% in the best cases as evident in Fig.~\ref{Fig:swhyc}.

\section {Conclusion and Future Work}
Here we have studied the important problem of automatically task-parallelizing ML dataflow graphs where batch size is 1 or a small positive number. Our goal has been to create fast algorithms that perform well. We devise linear clustering-based mechanisms to extract task parallelism from models augmented by advanced compiler optimizations like cloning and DCE. Our tool, Ramiel, converts ONNX models into parallel Pytorch+Python and demonstrates the characteristics of readability, executability and extendability via further optimizations such as intra-op/model/pipeline parallelism. New strategies like hyperclustering may turn out to be beneficial in training setups too. Finally, comparison with recent literature demonstrates the competitiveness of our approach. We have demonstrated that we can achieve fast compile time while obtaining similar performance as evidenced in (Table~\ref{tab:comp}). Future work involves further experimentation in power and resource-constrained settings as well as devising more powerful optimizations for graph reductions, possibly via tools like MLIR.

\bibliographystyle{IEEEtran}
\bibliography{ramielref}
\end{document}